\documentclass[sigconf, table]{acmart}
\usepackage{pifont}

\AtBeginDocument{%
  \providecommand\BibTeX{{%
    \normalfont B\kern-0.5em{\scshape i\kern-0.25em b}\kern-0.8em\TeX}}}

\setcopyright{acmlicensed}
\copyrightyear{2025}
\acmYear{2025}
\acmDOI{10.1145/3746027.3755700}

\acmConference[MM '25] {Proceedings of the 33rd ACM International Conference on Multimedia}{October 27--31, 2025}{Dublin, Ireland.}
\acmBooktitle{Proceedings of the 33rd ACM International Conference on Multimedia (MM '25), October 27--31, 2025, Dublin, Ireland}

\acmISBN{979-8-4007-2035-2/2025/10}

\begin{document}

\title{LMME3DHF: Benchmarking and Evaluating Multimodal 3D Human Face Generation with LMMs}


\author{Woo Yi Yang}
\affiliation{%
  \institution{Shanghai Jiao Tong University}
  \city{Shanghai}
  \country{China}}
\email{wooyiyang@sjtu.edu.cn}

\author{Jiarui Wang}
\affiliation{%
  \institution{Shanghai Jiao Tong University}
  \city{Shanghai}
  \country{China}}
\email{wangjiarui@sjtu.edu.cn}

\author{Sijing Wu}
\affiliation{%
  \institution{Shanghai Jiao Tong University}
  \city{Shanghai}
  \country{China}}
\email{wusijing@sjtu.edu.cn}

\author{Huiyu Duan}
\affiliation{%
  \institution{Shanghai Jiao Tong University}
  \city{Shanghai}
  \country{China}}
\email{huiyuduan@sjtu.edu.cn}

\author{Yuxin Zhu}
\affiliation{%
  \institution{Shanghai Jiao Tong University}
  \city{Shanghai}
  \country{China}}
\email{rye2000@sjtu.edu.cn}

\author{Liu Yang}
\affiliation{%
  \institution{Shanghai Jiao Tong University}
  \city{Shanghai}
  \country{China}}
\email{ylyl.yl@sjtu.edu.cn}

\author{Kang Fu}
\affiliation{%
  \institution{Shanghai Jiao Tong University}
  \city{Shanghai}
  \country{China}}
\email{fuk20-20@sjtu.edu.cn}

\author{Xiongkuo Min}
\authornote{Corresponding author.}
\affiliation{%
  \institution{Shanghai Jiao Tong University}
  \city{Shanghai}
  \country{China}}
\email{minxiongkuo@sjtu.edu.cn}

\author{Guangtao Zhai}
\affiliation{%
  \institution{Shanghai Jiao Tong University}
  \city{Shanghai}
  \country{China}}
\email{zhaiguangtao@sjtu.edu.cn}

\renewcommand{\shortauthors}{Woo Yi Yang and JiaRui Wang, et al.}

\begin{abstract}
The rapid advancement in generative artificial intelligence have enabled the creation of 3D human faces (HFs) for applications including media production, virtual reality, security, healthcare, and game development, \textit{etc}. However, assessing the quality and realism of these AI-generated 3D human faces remains a significant challenge due to the subjective nature of human perception and innate perceptual sensitivity to facial features. To this end, we conduct a comprehensive study on the quality assessment of AI-generated 3D human faces. We first introduce \textbf{Gen3DHF}, a large-scale benchmark comprising 2,000 videos of AI-\underline{Gen}erated \underline{3D} \underline{H}uman \underline{F}aces along with 4,000 Mean Opinion Scores (MOS) collected across two dimensions, \textit{i.e.}, quality and authenticity, 2,000 distortion-aware saliency maps and distortion descriptions. Based on Gen3DHF, we propose \textbf{LMME3DHF}, a \underline{L}arge \underline{M}ultimodal \underline{M}odel (LMM)-based metric for \underline{E}valuating \underline{3DHF} capable of quality and authenticity score prediction, distortion-aware visual question answering, and distortion-aware saliency prediction. Experimental results show that LMME3DHF achieves state-of-the-art performance, surpassing existing methods in both accurately predicting quality scores for AI-generated 3D human faces and effectively identifying distortion-aware salient regions and distortion types, while maintaining strong alignment with human perceptual judgments. Both the Gen3DHF database and the LMME3DHF will be released upon the publication.
\end{abstract}



\begin{teaserfigure}
  \centering
\includegraphics[width=1\linewidth]{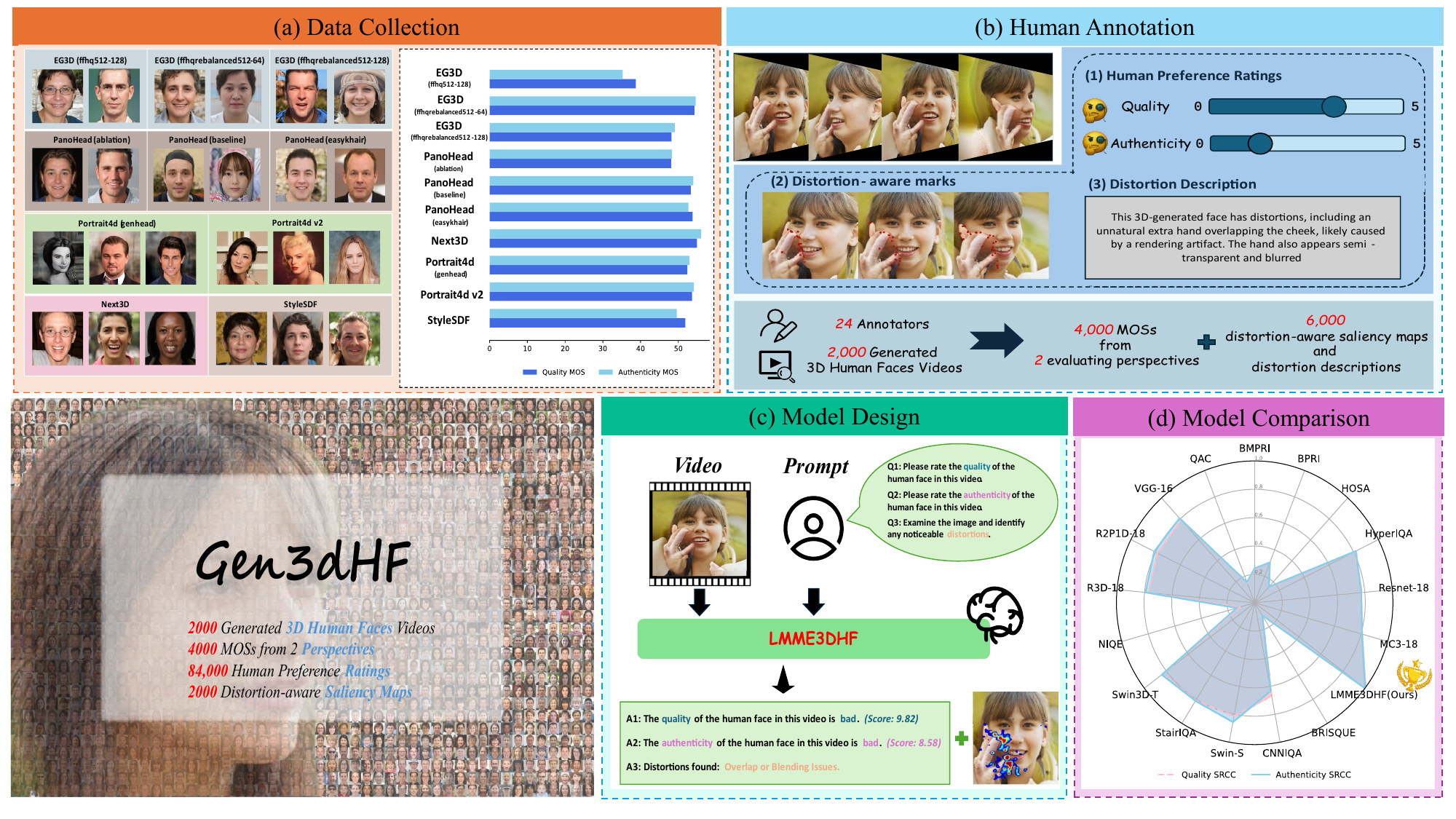}
\vspace{-8mm}
\caption{We present the generated 3D Human Face (HF) evaluation database and model, termed Gen3DHF and LMME3DHF, respectively.
(a) We first generate 2000 3D human faces using 5 3D human face generation models.
(b) 4000 MOS, 2000 distortion-aware saliency maps and distortion descriptions are acquired from 24 annotators.
(c) We design LMME3DHF to evaluate Gen3DHF.
(d) We compare the proposed LMME3DHF with other 16 benchmark models on our Gen3DHF dataset.}
\label{overview}
\vspace{2mm}
\end{teaserfigure}


\maketitle

\section{Introduction}
As digital communication continues to expand, conveying nuanced human attributes such as tone, emotion, and personality has become increasingly important. To address these challenges, digital humans powered by pioneering Generative Adversarial Networks (GANs) \cite{chan2022efficient, kammoun2022generative, skorokhodov2022epigraf} and recent diffusion models \cite{huang2024humannorm, stypulkowski2024diffused, huang2023collaborative}, have emerged as a promising solution. Among them, AI-generated 3D human faces (3D HFs) have gained particular attention for their role in enabling realistic avatars for applications in media production, virtual reality, gaming, and telepresence. Despite significant improvements in generation capabilities, state-of-the-art 3D HF generation models may still produce outputs suffering from perceptual distortions and unrealistic artifacts, failing to meet human quality expectations \cite{chan2022efficient, an2023panohead, sun2023next3d}. While human evaluation provides valuable insights, it remains prohibitively expensive and inefficient for large-scale assessment. Therefore, developing an objective quality metric that can accurately reflect human perception and preference for AI-generated 3D human faces is essential. However, existing quality assessment methods fall short in evaluating AI-generated 3D human faces due to facial distortions being inherently different from those in general AI-generated images or common objects.

In recent years, research on quality assessment of AI-generated content (AIGC) has gained significant momentum, several datasets have been proposed for image quality assessment (IQA) \cite{wang2023aigciqa2023, yang2024aigcoiqa2024} and video quality assessments (VQA) \cite{chivileva2023measuring, liu2024evalcrafter, liu2023fetv, zhang2024benchmarking, kou2024subjective, chen2024gaia}. Despite their contributions, these datasets are primarily designed for general objects and scenes, and thus not well-suited for evaluating AI-generated 3D human faces which present unique distortion patterns and none of the them are explicitly designed for 3D HF quality assessment. Traditional quantitative metrics such as Inception Score (IS) \cite{salimans2016improved} and Fréchet Inception Distance (FID) \cite{heusel2017gans} provide useful insights into overall model performance but exhibit fundamental limitations in evaluating the perceptual authenticity of individual generated samples. Traditional IQA methods \cite{kang2014convolutional, mittal2012making, su2020blindly, sun2023blind, xue2013learning}, while effective for evaluating individual natural images with common distortions \cite{duan2024finevq, min2019quality, zhai2020perceptual}, ignore the unique distortions in AI-generated HFs. Similarly, existing VQA methods \cite{li2022blindly, li2019quality, sun2022deep, wu2022fast, wu2023exploring} overlook the specialized requirements of 3D face evaluation. Moreover, these works only focus on quality assessment while overlooking the critical need for distortion region localization. Traditional saliency prediction methods \cite{bruce2005saliency, goferman2011context, erdem2013visual}, which simply identify visually prominent regions, fail to distinguish between naturally salient facial areas and those containing significant quality degradations.  

In this paper, we introduce Gen3DHF, a comprehensive dataset and benchmark comprising 2,000 diverse 3D HF video samples generated by five distinct models. As illustrated in Figure \ref{overview}, we collect 90,000 human annotations, resulting in 4,000 Mean Opinion Scores (MOS) and 2,000 distortion-aware saliency maps with corresponding distortion descriptions. Based on Gen3DHF, we propose LMME3DHF, a LMM-based metric specifically designed to not only evaluate 3D HF content across the two dimensions, \textit{i.e.}, quality and authenticity, but also to predict and output the salient distortion regions along with their corresponding textual descriptions. LMME3DHF leverages instruction tuning \cite{liu2023visual} and LoRA adaptation\cite{hu2022lora} techniques to fine-tune the language model. 

Our main contributions are summarized as follows:
\begin{itemize}
    \item We construct Gen3DHF, a comprehensive dataset consisting of 2,000 diverse AI-generated 3D HF videos produced by five different models, annotated with 4,000 Mean Opinion Scores (MOSs) from perspectives of quality and authenticity dimensions, and 2,000 distortion-aware saliency maps with corresponding distortion descriptions.
    \item We propose LMME3DHF, a LMM-based metric capable of providing detailed text-based quality level evaluations, and precise quantitative predictions of quality scores for 3D HF.
    \item LMME3DHF, equipped with a saliency decoder, can accurately predict distortion-aware saliency and identify the corresponding distortion type.
    \item Extensive experiments demonstrate LMME3DHF's state-of-the-art performance in both accurately predicting quality scores for AI-generated 3D human faces and effectively identifying distortion-aware salient regions and distortion types.
\end{itemize}
\begin{figure*}[t]
    \centering
    \begin{minipage}[t]{0.49\textwidth}
        \centering
        \includegraphics[width=\linewidth]{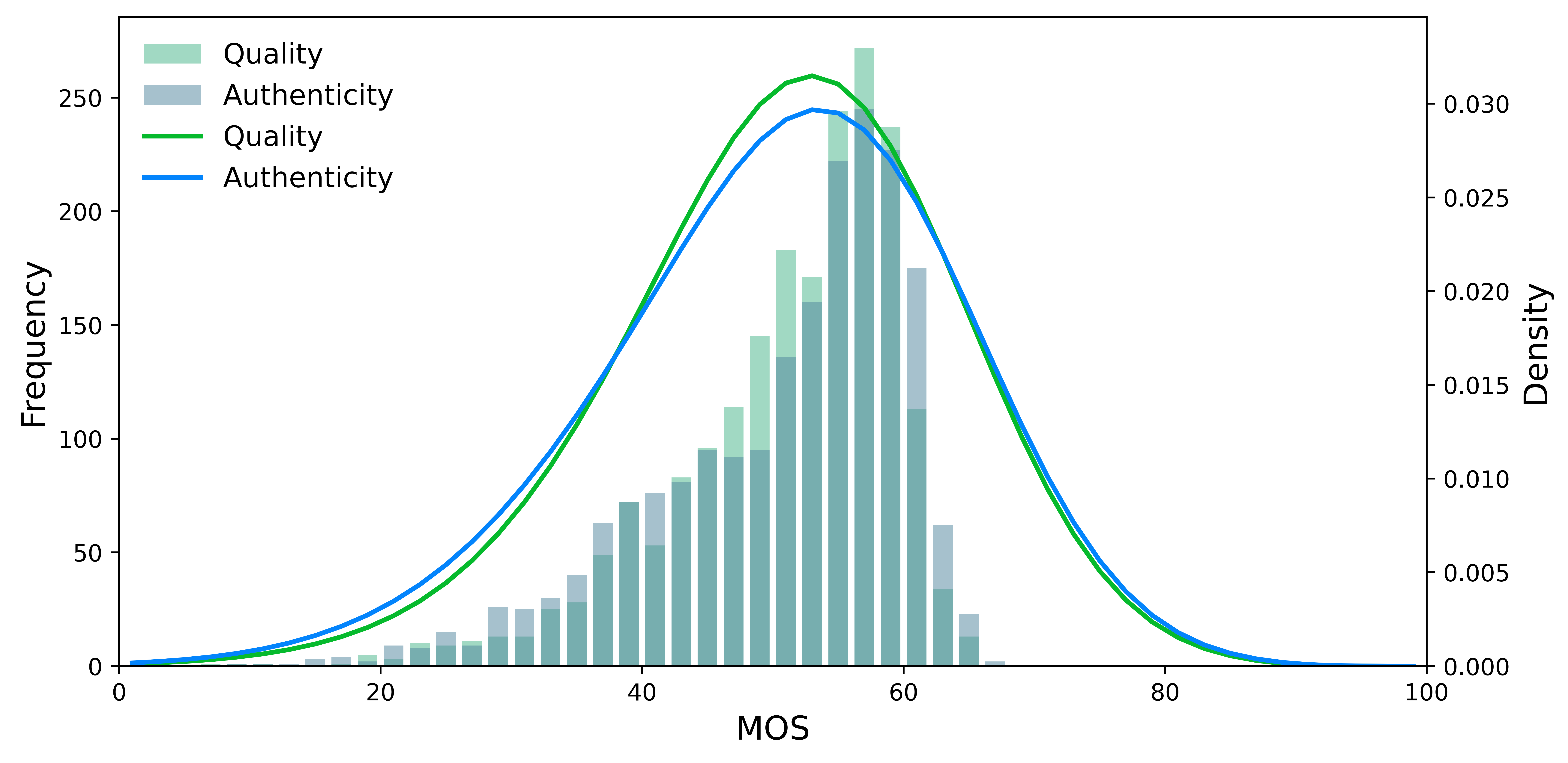}
        (a)
    \end{minipage}
    \hfill
    \begin{minipage}[t]{0.49\textwidth}
        \centering
        \includegraphics[width=\linewidth]{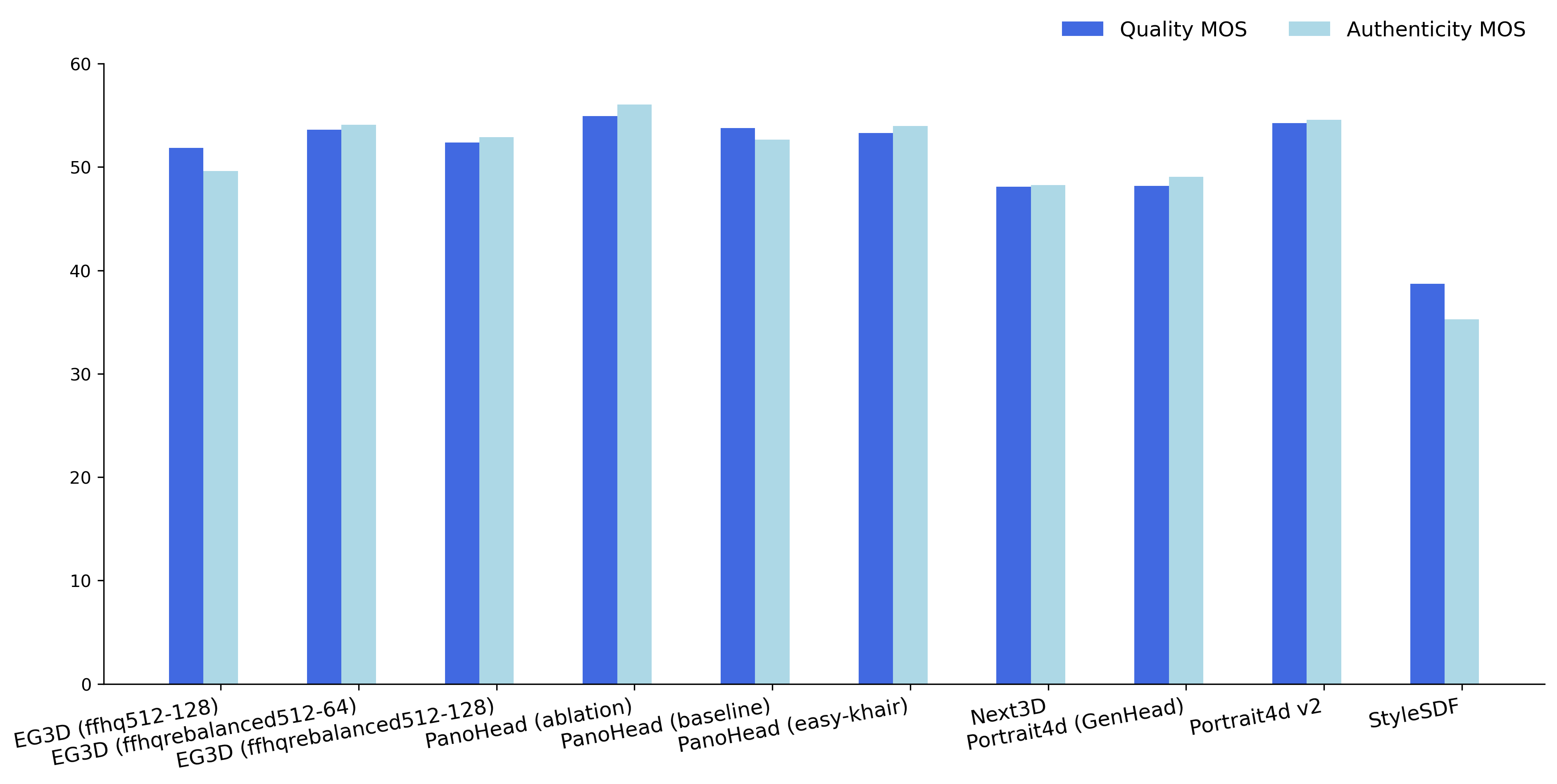}
        (b)
    \end{minipage}
    \vspace{-4mm}
    \caption{(a) Distribution of quality and authenticity MOSs; (b) Comparison of 3D HF generation models regarding the quality MOSs and authenticity MOSs.}
    \vspace{-2mm}
    \label{MOS}
\end{figure*}

\section{RELATED WORK}
\subsection{3D Human Face Generation}
Recent advancements in 3D-aware generative adversarial networks (GANs) significantly improve the quality of synthesized human faces, achieving breakthroughs in photorealism, multi-view consistency, and fine-grained controllability. While explicit and implicit neural rendering methods each have limitations in scalability or fidelity, \cite{chan2022efficient} introduced a hybrid explicit–implicit tri-plane architecture that achieves high-resolution, multi-view-consistent synthesis from unstructured 2D images. Building on it, \cite{an2023panohead, sun2023next3d, wu2022anifacegan, wu2023aniportraitgan, deng2024portrait4d, deng2024portrait4dv2, or2022stylesdf, schwarz2022voxgraf, zhang2022multi, sun2022controllable} extend 3D HF generation to full \begin{math}360^\circ\end{math} head views, introduce animatable controls over facial expressions and head–shoulder movements respectively, detailed facial deformation with high fidelity, demonstrated high-fidelity 4D head synthesis and motion transfer using synthetic or pseudo multi-view data, achieving robust motion control without relying on monocular 3D Morphable Model (3DMM) reconstruction. Nevertheless, 3D human face generation still suffers from certain distortions, such as inconsistent views across frames, unrealistic geometry in unseen regions, or artifacts introduced by limited training data and view ambiguity.

\subsection{Video Quality Assessment}
With the rapid advancement of AI-generated image (AIGI), various image quality assessment (IQA) methods \cite{li2023agiqa, wang2023aigciqa2023, liang2024rich, kirstain2023pick, wu2023better} have emerged to evaluate the perceptual quality of these images. As the capabilities of AI-generated content continue to evolve \cite{dhariwal2021diffusion, rombach2022high}, AI-generated videos (AGVs) are now receiving increasing attention and being applied across a wide range of use cases. To assess their quality, several video quality assessment (VQA) studies have been conducted \cite{chen2024gaia, chivileva2023measuring, liu2024evalcrafter, liu2023fetv, kou2024subjective, huang2024vbench}. Early VQA methods often adopted evaluation metrics originally designed for IQA. While effective for measuring basic perceptual similarity, these metrics do not fully capture temporal coherence or motion consistency in videos. More recently, Contrastive Language–Image Pre-training (CLIP) \cite{radford2021learning} has been employed as a backbone for quality assessment tasks, leveraging its powerful multimodal representations to better align generated content with human-perceived semantics. Meanwhile, the rise of video large multimodal models introduces semantic-aware quality assessment, where models pre-trained on vision-language tasks jointly analyze visual fidelity and textual descriptions and particularly excel in visual question-answering. Existing deep learning-based VQA methods lack interactive QA capabilities, while video large multimodal models show limited precision in perceptual quality assessment.

\subsection{Saliency Prediction}
Recent advances in computer vision have driven significant progress in human visual attention analysis, with saliency prediction emerging as a crucial research area. This technology simulates human visual system's mechanisms for allocating attention across different image regions. In recent years, numerous research efforts have focused on developing saliency prediction models. Traditional models \cite{bruce2005saliency, goferman2011context, erdem2013visual} primarily rely on low-level features such as intensity, color, and contrast to generate saliency maps. In contrast, deep learning-based models \cite{huang2015salicon, che2019gaze, lou2022transalnet} are capable of capturing more complex visual patterns and semantic information. However, despite their success in general applications, these models often fall short in the context of quality assessment, where identifying distortion-specific regions is crucial for accurately evaluating perceptual quality. This motivates us to develop a saliency prediction model specifically tailored for identifying distortion-aware salient regions in AI-generated human faces as distortion-aware saliency prediction not only guides model attention toward visually critical regions but also enhances interpretability by highlighting the exact areas that degrade user experience.

\begin{figure*}[t]
    \centering
    \begin{minipage}[t]{0.49\textwidth}
        \centering
        \includegraphics[width=\linewidth]{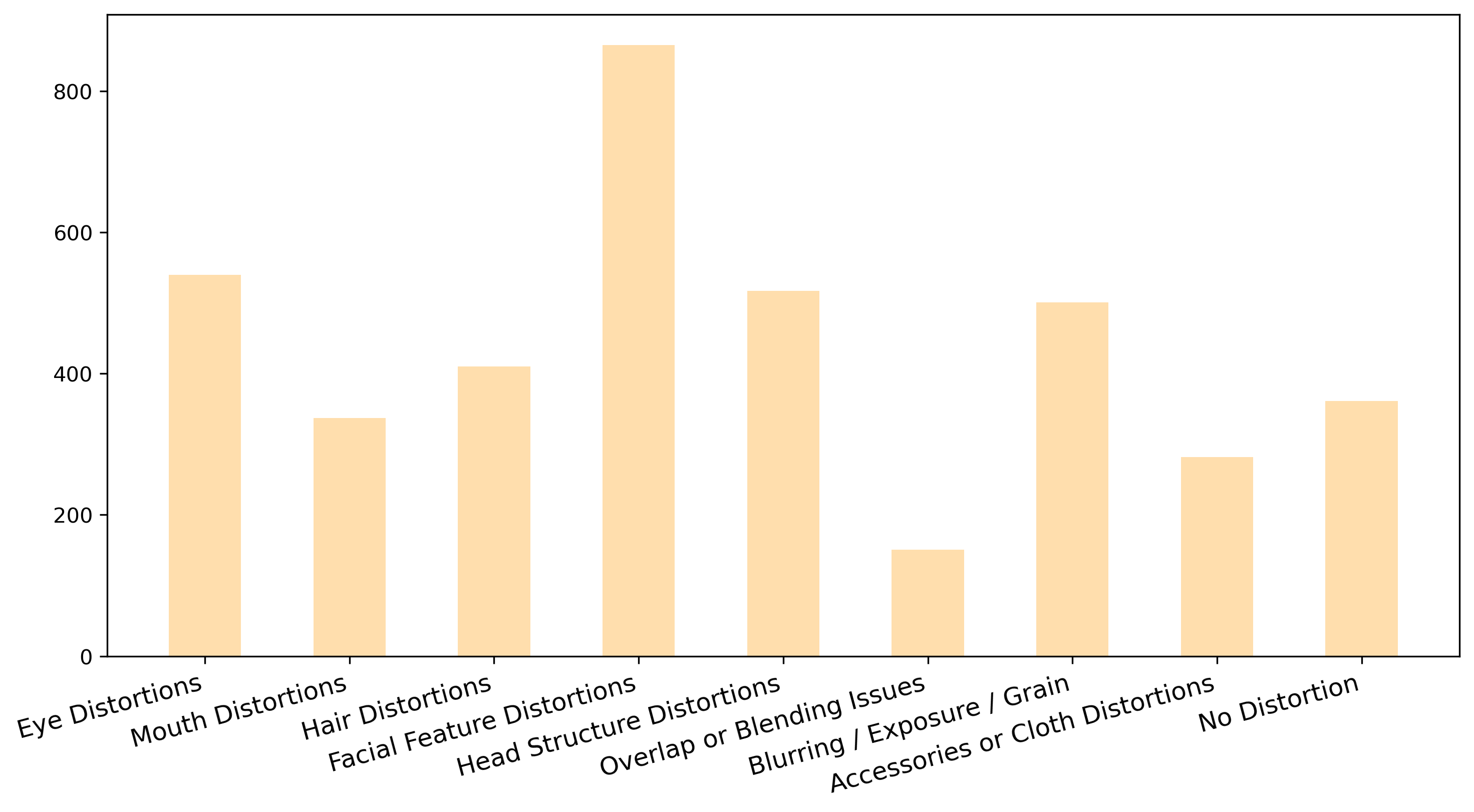}
        \raisebox{6mm}{(a)}
    \end{minipage}
    \hfill
    \begin{minipage}[t]{0.49\textwidth}
        \centering
        \includegraphics[width=\linewidth]{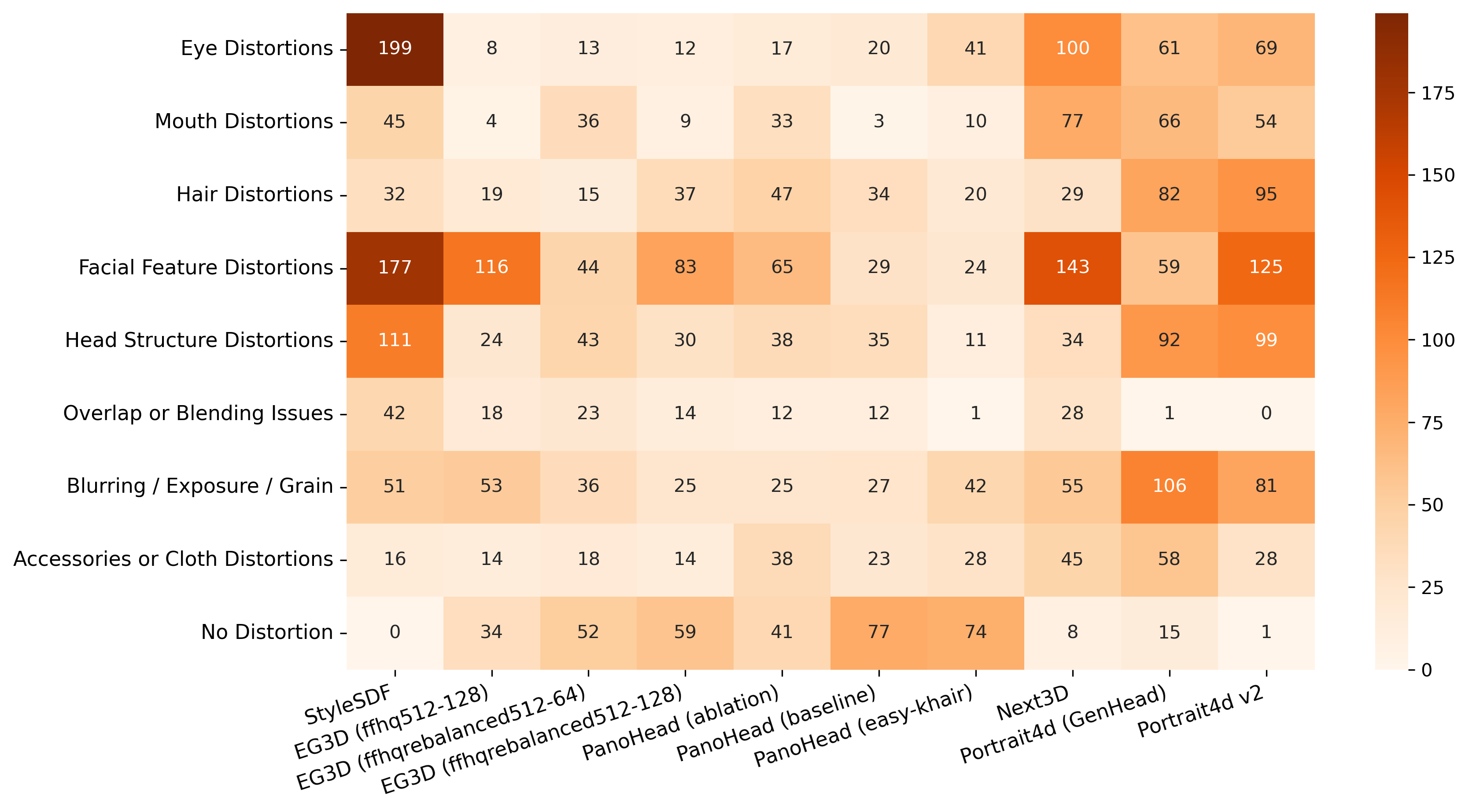}
        \raisebox{6mm}{(b)}
    \end{minipage}
    \vspace{-8mm}
    \caption{(a) Distribution of distortion categories; (b) Heatmap showing the frequency of different distortion types detected across 10 generation models.}
    \vspace{-4mm}
    \label{distortion}
\end{figure*}

\begin{figure}[t]
    \centering
    \includegraphics[width=\linewidth]{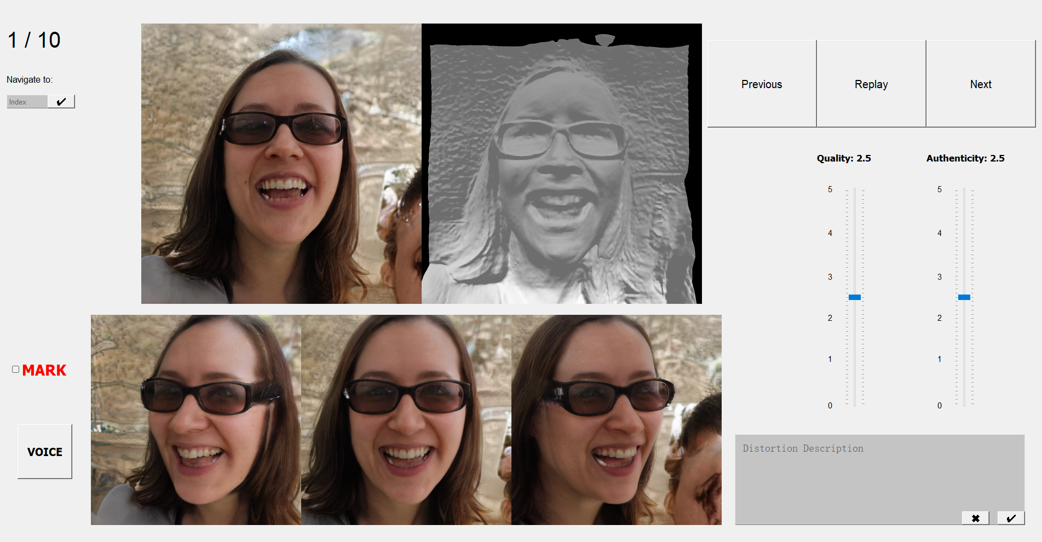}
     \vspace{-6mm}
    \caption{The subjective experiment user interface.} 
     \vspace{-8mm}
    \label{interface}
\end{figure}

\section{DATABASE CONSTRUCTION}
\subsection{3D Human Face Generation}
To ensure content diversity, we utilize five different 3D Human Face (HF) content generation models, including \cite{chan2022efficient, an2023panohead, sun2023next3d, deng2024portrait4dv2, or2022stylesdf} to produce 3D HF using open source code and default weights. Each sample consists of paired 512 \begin{math}\times\end{math} 512 resolution RGB videos and 3D mesh sequences, rendered using a standardized camera setup that performs a uniformly rotation around the 3D HF generated to fully capture frontal facial features. All videos are rendered at 60 frames per second (FPS) using the same camera setup, which performs a \begin{math}180^\circ\end{math} rotation to capture the complete frontal view of the HF. Each content generated is four seconds long. Therefore, the Gen3DHF resulting in a total of 2,000 diverse instances (200 seeds \begin{math}\times\end{math} 10 pre-trained networks) of 3D HF content.

\subsection{Subjective Experiment}
We propose a dual-perspective evaluation framework to better align VQA with human preferences. The quality dimension assesses perceptual attributes such as texture, color fidelity, and structural integrity. In contrast, the authenticity dimension evaluates perceptual aspects like unidentified object appearance, head structure and the presence of unrealistic textures or shapes. 

Subjective experiments are conducted in accordance with the ITU-R BT.500-13 \cite{series2012methodology} guidelines. As shown in Figure \ref{interface}, the interface presents a video along with a static image containing three snapshots from the video at \begin{math}-45^\circ\end{math}, \begin{math}0^\circ\end{math} and \begin{math}45^\circ\end{math} angles (with \begin{math}0^\circ\end{math} representing the frontal view). Participants can navigate the 3D HF projection videos using ``\verb|Previous|'', ``\verb|Replay|'' and ``\verb|Next|'' interactive buttons. Additionally, two sliders ranging from 0 to 5 enable participants to score the content on both quality and authenticity dimensions smoothly. A ``\verb|MARK|'' button is provided to allow participants to highlight any visible distortions on the static images. If distortions are marked, participants are instructed to provide descriptions by typing or voice recording through ``\verb|Voice|'' button. A total of 24 subjects participate in the subjective experiment. All participants receive a thorough briefing beforehand. Each session lasts approximately two hours. Finally, we obtain a total of 90,000 human annotations including 84,000 reliable score ratings (21 annotators \begin{math}\times\end{math} 2 dimensions \begin{math}\times\end{math} 2,000 videos), 6,000 distortion-aware saliency maps and distortion descriptions (3 annotators \begin{math}\times\end{math} 2,000 images) which were then aggregated into a single combined annotation per image.

\begin{figure}[t]
    \centering
    \includegraphics[width=\linewidth]{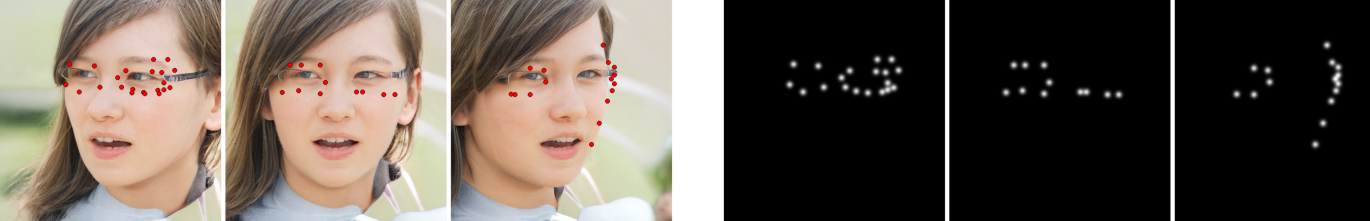}
     \vspace{-4mm}
    \makebox[\linewidth]{\hspace{0.3cm}(a) \hspace{4cm} (b)}
    \caption{(a) Images with human-annotated distortion points; (b) Generated distortion-aware aliency maps.} 
     \vspace{-4mm}
    \label{saliency_preprocess}
\end{figure}
\subsection{Subjective Data Processing}
By following the suggestion recommended by ITU to conduct the outlier detection and subject rejection. We first convert the raw ratings into Z-scores, which were then linearly scaled to the range of [0,100] as follows:
\begin{equation}
    z_{ij} = \frac{r_{ij} - \mu_{ij}}{\sigma_I} \,, \quad
    z_{ij}' = \frac{100 \times(z_{ij} + 3)}{6}
\end{equation}
\begin{equation}
    \mu_i = \frac{1}{N} \sum_{j=1}^{N_i} r_{ij} \,, \quad
    \sigma_i = \sqrt{\frac{1}{N_{i}-1} \sum_{j=1}^{N_i} (r_{ij} - \mu_{ij})^2}
\end{equation}
\begin{equation}
    MOS_j = \frac{1}{N} z_{ij}'
\end{equation}
where \begin{math}r_{ij}\end{math} is the raw rating given by the \begin{math}i\end{math}-th subject to the \begin{math}j\end{math}-th videos. \begin{math}N_i\end{math} is the number of videos judged by subject \begin{math}i\end{math}. The MOS of the \begin{math}j\end{math}-th videos is then computed as Equation (3) where \begin{math}MOS_{j}\end{math} indicates the MOS for the \begin{math}j\end{math}-th 3D HF. Therefore, there are total 4000 MOSs (2 dimensions \begin{math}\times\end{math} 2000 videos) obtained.

Additionally, to quantify performance and standardize raw distortion descriptions, we classify distortions into nine distinct categories based on their visual characteristics \begin{math}d_i\end{math}, where \begin{math}\{d_i|_{i=1}^{9} \} = \end{math} \{\textit{"Eye Distortions", "Mouth Distortions", "Hair Distortions", "Facial Feature Distortions", "Head Structure Distortions", "Overlap or Blending Issues", "Blurring / Exposure / Grain", "Accessories or Cloth Distortions”, “No Distortion”}\}. Due to the nature of our subjective experiment data where distortion regions are mark using red dots, a transformation is required to convert these discrete annotations into continuous distortion-aware saliency maps. As illustrated in Figure \ref{saliency_preprocess}, and following standard practices in saliency prediction tasks, we first identify the center of each distorted region based on the human-marked red dots. These fixation maps are then smoothed using a Gaussian filter with a standard deviation of \begin{math}\sigma = 5\end{math}.

\begin{figure}[t]
    \centering
    \includegraphics[width=\linewidth]{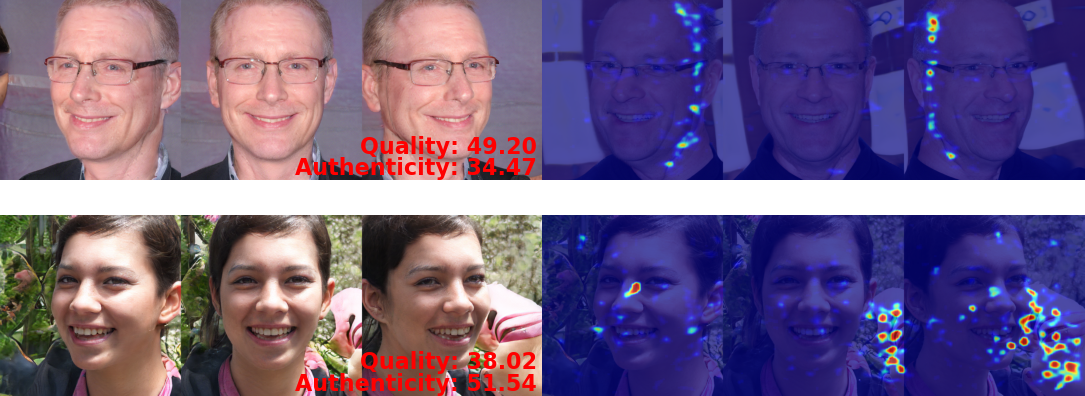}
     \vspace{-6mm}
    \caption{Examples of the human visual experience scores and distortion-aware saliency maps.} 
     \vspace{-4mm}
    \label{data_example}
\end{figure}
\subsection{Subjective Data Analysis}
The MOS and score distributions for the quality and authenticity dimensions are illustrated in Figure \ref{MOS}(a). Both distributions are slightly right-skewed, indicating that most samples received moderate to high ratings, peaking around the 55–60 range. This suggests that the majority of the generated 3D human face (HF) content was perceived as reasonably convincing in terms of both visual quality and authenticity. To further investigate the perceptual performance of different 3D HF generation models, we compare their MOS for quality and authenticity. As shown in Figure \ref{MOS}(b), EG3D \cite{chan2022efficient} performs well in terms of quality but lags in authenticity, whereas PanoHead \cite{an2023panohead} exhibits the opposite trend. This contrast underscores the importance of evaluating quality and authenticity as separate dimensions. As illustrated in Figure \ref{distortion}(a), each distortion type occurs regularly, with facial feature distortions appearing most frequently. This indicates that AI-generated 3D HFs particular challenges in generating smooth, realistic, and detailed facial features. The distribution of distortion types across different generation models is illustrated in Figure \ref{distortion}(b). As shown in Figure \ref{data_example}, we observe that an increased number of distortion-aware salient regions strongly correlates with a significantly worse human visual experience. In other words, lower scores are associated with more extensive and concentrated saliency in distorted areas.

\begin{figure*}
    \centering
    \vspace{-5mm}
    \includegraphics[width=1\textwidth]{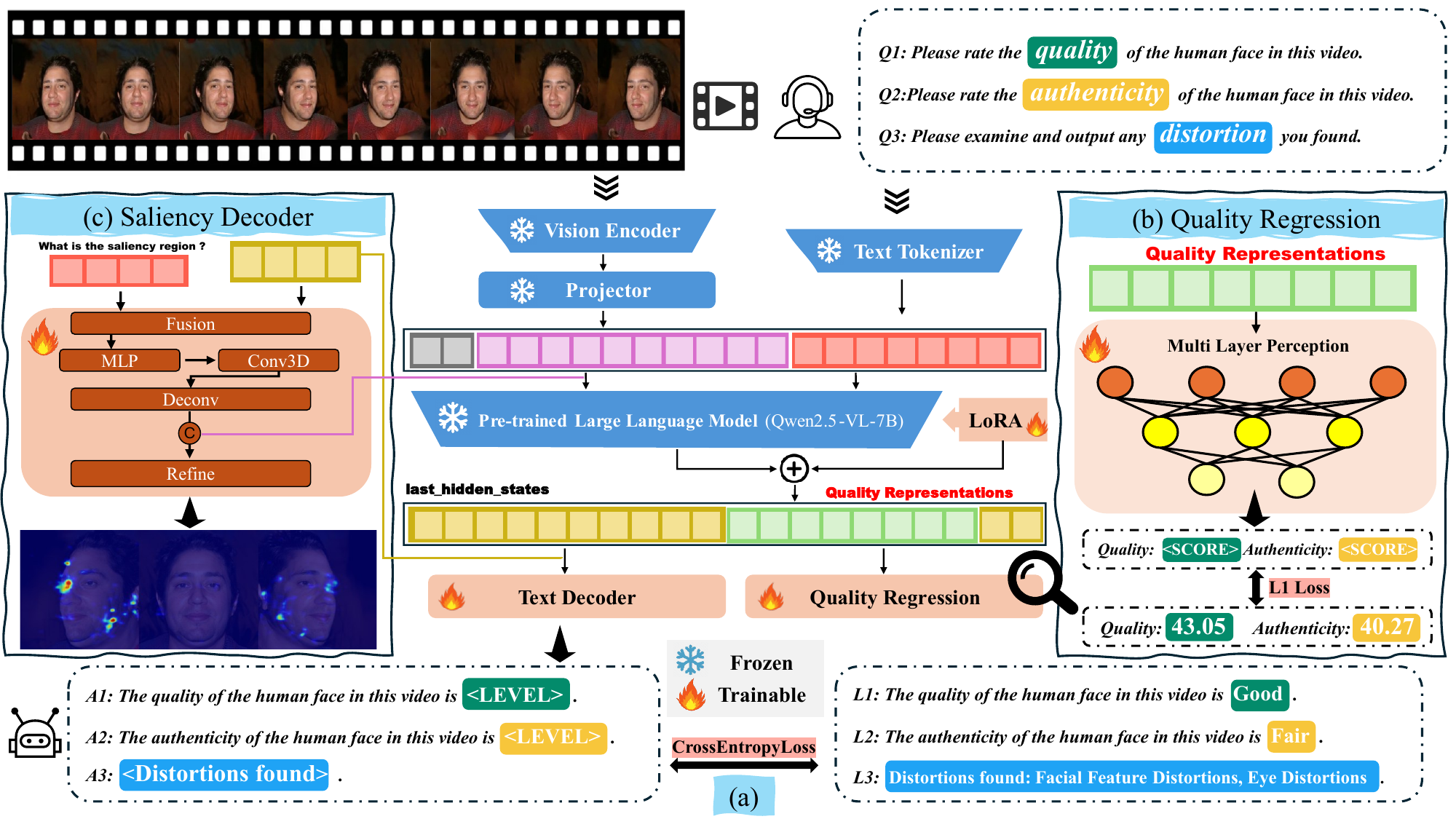}
    \vspace{-6mm}
    \caption{Overview of the LMME3DHF architecture. The model includes three functions: (a) text-defined quality level prediction, (b) precises score prediction, (c) distortion-aware saliency map prediction. The training process consists of three stages: (a) instruction tuning of the model via text-defined levels and distortion description, (b) fine-tuning the LMM via numerical scores, (c) train saliency decoder from scratch.} 
     \vspace{-1mm}
    \label{model}
\end{figure*}

\section{PROPOSED MODEL}
In this section, we introduce our \textbf{\textit{all-in-one}} 3D HFs quality assessment framework, LMME3DHF, which is designed to generate text-defined quality level descriptions, predict precise quality and authenticity scores, and produce distortion-aware saliency maps along with corresponding textual descriptions.

\subsection{Model Structure}
\noindent\textbf{Visual Encoding.} \quad As shown in Figure \ref{model}, we leverage the pretrained Vision Transformer (ViT) module embedded in Qwen2.5-VL-7B to extract visual features. This model demonstrates strong capabilities in handling videos with varying resolutions and frame rates, allowing us to bypass preprocessing steps such as compression or resizing, which often lead to information loss. The native projector, implemented using two-layer Multi-Layer Perceptron (MLP), is used to project the extracted visual features into a latent space aligned with the text embeddings used in the language model.

\noindent\textbf{Quality Assessment.} \quad We further utilize the LMM (Qwen2.5-VL-7B) to integrate visual and textual information for quality assessment. Our quality evaluation framework is divided into two subtasks: (1) Qualitative description generation: The model generates a descriptive textual assessment of the video’s quality, such as: “The quality of the human face in this video is (bad, poor, fair, good, excellent).” Since LMM excel at interpreting and generating textual information, this task provides intuitive and user-friendly feedback. Moreover, it serves as a form of preliminary classification that helps guide the subsequent regression task. (2) Quantitative score regression: The model uses the final hidden states from the LMM to perform a regression task, predicting numerical scores for both quality and authenticity dimensions. This dual-output design ensures that the model can deliver both interpretive and quantitative assessments of AI-generated 3D HF content.

\noindent\textbf{Distortion-aware Saliency Decoder.} \quad The distortion-aware saliency decoder is built upon a U-Net architecture, leveraging both the visual features extracted by the vision encoder and multimodal features from language decoder to generate final saliency map. The visual features are processed through a series of convolutional layers and PixelShuffle operations to enhance spatial resolution. Simultaneously, the multimodal features pass through MLP followed by 3D convolutional layers to capture spatiotemporal context. These processed features are then fused and refined using spatial convolutional layers and PixelShuffle operations to recover fine-grained detail. Finally, a set of convolutional layers produces the output saliency map, effectively highlighting regions of visual distortion within the 3D human face video content.

\subsection{Training and Fine-tuning Strategy}
The training process of LMME3DHF follows a three-stage approach designed to ensure robust video assessment capabilities across three core tasks: quality level prediction, individual quality scoring, and distortion-aware saliency prediction. The three stages are: (1) Instruction tuning using text-defined quality levels and distortion categories, (2) Fine-tuning the LMM with LoRA for precise quality score regression, and (3) Training the saliency decoder from scratch to generate distortion-aware saliency maps.

\noindent\textbf{Instruction Tuning.} \quad Due to the inherent strengths of LLMs in processing textual rather than numerical data, we begin by converting quality scores into categorical, text-based quality levels. We uniformly divides the score range from the minimum value (m) to the maximum value (M) into five equal intervals and assigns each interval a corresponding textual label based on Equation (4).
\begin{equation}
    L(s) = l_{i} \quad if \quad m + \frac{i-1}{5} \times (M-m) < s \leq m +\frac{1}{5} \times (M-m) 
\end{equation}
where \begin{math}\{l_i |_{i=1}^{5} \} = \{bad,poor,fair,good,excellent\}\end{math}. The distortion descriptions were categorized into nine classes, as previous literature described. During instruction tuning, the model is trained using language loss, guiding it to generate quality-level predictions and distortion category classifications.

\noindent\textbf{Quality Regression Fine-Tuning.} \quad To enable continuous quality score prediction, we implement a quality regression module. This module takes the last hidden states from the LLM as input and outputs precise predicted scores. As full fine-tuning of LLMs can be computationally expensive, we leverage the LoRA \cite{hu2022lora} technique to efficiently adapt the model to the quality regression task. During this fine-tuning phase, we leverage L1 loss to align predicted scores with ground-truth MOS.

\noindent\textbf{Distortion-aware saliency maps.} \quad To produce distortion-aware saliency maps, we introduce a saliency decoder that integrates visual features from the vision encoder and multimodal features from the language decoder. The decoder processes these inputs to generate a saliency map that highlights regions of distortion within the video content. The training loss function for the saliency decoder is defined in Equation (5), which is a linear combination of four loss functions: L1 Loss, Correlation Coefficient Loss, KL Divergence Loss, and Binary Cross-Entropy Loss. 
\begin{equation}
    \mathcal{L} = \omega_{1}\mathcal{L}_{L1} + \omega_{2}\mathcal{L}_{CC} + \omega_{3}\mathcal{L}_{KL} + \omega_{4}\mathcal{L}_{BCE}   
\end{equation}

\section{EXPERIMENTS}
\subsection{Experiment Setup}
\begin{table}[t]
\vspace{-3mm}
\caption{Performance comparisons of the state-of-the-art quality evaluation methods on the Gen3DHF from perspectives of Quality and Authenticity. $\spadesuit$ Handcrafted IQA models, $\diamondsuit$ Deep learning-based IQA models, $\clubsuit$ Deep learning-based VQA models,  $\heartsuit$ LMM-based models. The best results are marked in \textcolor{red}{RED} and the second-best in \textcolor{blue}{BLUE}.
}
\centering
\vspace{-3mm}
\resizebox{0.47\textwidth}{!}{
  \begin{tabular}{l||ccc|ccc}
    \toprule

    \multicolumn{1}{l}{\bf Dimension} &
    \multicolumn{3}{c}{\textbf{Quality}} &
    \multicolumn{3}{c}{\textbf{Authenticity}} 
    \\
    \cmidrule(lr){2-4} \cmidrule(lr){5-7}
    \textbf{Methods / Metrics}  
    & \textbf{SRCC}  & \textbf{PLCC}  & \textbf{KRCC}
    & \textbf{SRCC}  & \textbf{PLCC}  & \textbf{KRCC}

\\
\hline
$\spadesuit$ \textbf{BMPRI} \cite{min2018blind}
 & 0.2344 & 0.2728 & 0.1591 
 & 0.2428 & 0.2892 & 0.1662

\\
$\spadesuit$ \textbf{BPRI} \cite{min2017blind} 
& 0.3069  & 0.3530 & 0.2100 
& 0.3050  & 0.3596 & 0.2093 

\\
$\spadesuit$ \textbf{BPRI-PSS} \cite{min2017blind} 
 & 0.2714 & 0.2998 & 0.1857 
 & 0.2711 & 0.3089 & 0.1848 

\\
$\spadesuit$ \textbf{BPRI-LSSs}  \cite{min2017blind}
 & 0.2741 & 0.3077 & 0.1864
 & 0.2736 & 0.3164 & 0.1870

\\
 
$\spadesuit$ \textbf{BPRI-LSSn} \cite{min2017blind} 
 & 0.1121 & 0.0999 & 0.0751
 & 0.1174 & 0.1168 & 0.0783

\\
$\spadesuit$ \textbf{BRISQUE} \cite{mittal2012no} 
 & 0.0989 & 0.0929 & 0.0664
 & 0.1035 & 0.1162 & 0.0679

\\
$\spadesuit$ \textbf{HOSA} \cite{xu2016blind}
 & 0.1479 & 0.1793 & 0.0994
 & 0.1511 & 0.1944 & 0.1019

\\
$\spadesuit$ \textbf{NIQE} \cite{mittal2012making}
 & 0.1107 & 0.1807 & 0.0732
 & 0.1336 & 0.2103 & 0.0896

\\
$\spadesuit$ \textbf{QAC} \cite{xue2013learning}
 & 0.1507 & 0.1858 & 0.1012
 & 0.1555 & 0.2102 & 0.1031

\\
\hline
$\diamondsuit$ \textbf{Resnet-18} \cite{he2016deep}
 & 0.7716 & 0.8087 & 0.5773
 & 0.7743 & 0.8173 & 0.5857

\\
$\diamondsuit$ \textbf{Resnet-34} \cite{he2016deep}
 & 0.7699 & 0.8041 & 0.5761
 & 0.7692 & 0.8054 & 0.5720

\\
$\diamondsuit$ \textbf{Resnet-50} \cite{he2016deep}
 & 0.7649 & 0.7886 & 0.5755
 & 0.7431 & 0.7752 & 0.5524

\\
$\diamondsuit$ \textbf{VGG-16} \cite{simonyan2014very}
 & 0.7891 & 0.8339 & 0.5982
 & 0.8064 & 0.8319 & 0.6122

\\
$\diamondsuit$ \textbf{VGG-19} \cite{simonyan2014very}
 & 0.7950 & 0.8379 & 0.6021
 & 0.8212 & 0.8514 & 0.6298

\\
$\diamondsuit$ \textbf{Swin-T} \cite{liu2021swin}
 & 0.8035 & 0.8476 & 0.6133
 & 0.8302 & 0.8638 & 0.6368

\\
$\diamondsuit$ \textbf{Swin-S} \cite{liu2021swin}
 & 0.8132 & 0.8318 & 0.6311
 & \bf\textcolor{blue}{0.8554} & 0.8589 & \bf\textcolor{blue}{0.6611}

\\
$\diamondsuit$ \textbf{Swin-B} \cite{liu2021swin}
 & 0.7512 & 0.8348 & 0.5688
 & 0.7945 & 0.8565 & 0.6131

\\
$\diamondsuit$ \textbf{Swin-L} \cite{liu2021swin}
 & 0.7961 & 0.8627 & 0.6168
 & 0.8235 & 0.8754 & 0.6450

\\
$\diamondsuit$ \textbf{CNNIQA} \cite{kang2014convolutional}
 & 0.6664 & 0.7027 & 0.4787
 & 0.6321 & 0.6724 & 0.4497

\\
$\diamondsuit$ \textbf{StairIQA} \cite{sun2023blind}
 & 0.8098 & 0.8491 & 0.6189
 & 0.8134 & 0.8376 & 0.6303

\\
$\diamondsuit$ \textbf{HyperIQA} \cite{su2020blindly}
& 0.8106 & 0.8473 & 0.6232
& 0.8164 & 0.8577 & 0.6289

\\
\hline
$\clubsuit$ \textbf{MC3-18} \cite{tran2018closer}
 & 0.8076 & 0.8464 & 0.6166
 & 0.8071 & 0.8439 & 0.6111

\\
$\clubsuit$ \textbf{R2P1D-18} \cite{tran2018closer}
 & 0.7865 & 0.8320 & 0.5941
 & 0.8113 & 0.8364 & 0.6137

\\
$\clubsuit$ \textbf{R3D-18} \cite{tran2018closer}
 & 0.7734 & 0.8102 & 0.5788
 & 0.7932 & 0.8255 & 0.5972

\\
$\clubsuit$ \textbf{Swin3D-T} \cite{liu2022video}
 & \bf\textcolor{blue}{0.8333} & \bf\textcolor{blue}{0.8706} & \bf\textcolor{blue}{0.6473}
 & 0.8413 & \bf\textcolor{blue}{0.8737} & 0.6516

\\
$\clubsuit$ \textbf{Swin3D-S} \cite{liu2022video}
 & 0.7978 & 0.8500 & 0.6064
 & 0.8243 & 0.8623 & 0.6342

\\
$\clubsuit$ \textbf{Swin3D-B} \cite{liu2022video}
 & 0.7922 & 0.8419 & 0.6000
 & 0.8321 & 0.8637 & 0.6381

\\
$\clubsuit$ \textbf{Simple-VQA} \cite{sun2022deep}
 & 0.6371 & 0.6753 & 0.4474
 & 0.5930 & 0.6714 & 0.4177

\\
$\clubsuit$ \textbf{Fast-VQA} \cite{wu2022fast}
 & 0.7538 & 0.7965 & 0.5645
 & 0.7451 & 0.7951 & 0.5551
\\
$\clubsuit$ \textbf{DOVER} \cite{wu2023exploring}
 & 0.6564 & 0.6676 & 0.4639
 & 0.6017 & 0.6815 & 0.4301
 
\\
$\clubsuit$ \textbf{TCSVT-BVQA} \cite{li2022blindly}
 & 0.7858 & 0.7985 & 0.5952
 & 0.7713 & 0.7924 & 0.5816
 
\\
$\clubsuit$ \textbf{VSFA} \cite{li2019quality}
 & 0.7501 & 0.8154 & 0.5671
 & 0.7305 & 0.8010 & 0.5483

\\
\hline
$\heartsuit$ \textbf{VideoChat2 (7B)} \cite{li2023videochat} 
 & 0.1959 & 0.2065 & 0.1367
 & 0.0596 & 0.1769 & 0.0427

\\
$\heartsuit$ \textbf{LLaVA-NeXT (8B)} \cite{li2024llava}
 & 0.0702 & 0.1435 & 0.0550
 & 0.0343 & 0.0414 & 0.0280

\\
$\heartsuit$ \textbf{InternVL2.5 (8B)} \cite{chen2024expanding}
 & 0.1226 & 0.0593 & 0.0935
 & 0.2026 & 0.2245 & 0.1585

\\
$\heartsuit$ \textbf{MiniCPM-V2.6 (8B)} \cite{yao2024minicpm}
 & 0.1331 & 0.1137 & 0.0972
 & 0.0989 & 0.1070 & 0.0710

\\
$\heartsuit$ \textbf{Qwen2-VL (7B)} \cite{wang2024qwen2}
 & 0.1329 & 0.1798 & 0.1079
 & 0.0719 & 0.1695 & 0.0558

\\
$\heartsuit$ \textbf{Video-ChatGPT} \cite{maaz2023video}
 & 0.0355 & 0.0528 & 0.0258
 & 0.0011 & 0.0123 & 0.0004

\\
$\heartsuit$ \textbf{Video-LLaVA} \cite{lin2023video}
 & 0.0586 & 0.0783 & 0.0477
 & 0.1199 & 0.1258 & 0.0978

\\
$\heartsuit$ \textbf{Video-LLaMA2} \cite{cheng2024videollama}
 & 0.1976 & 0.2806 & 0.1547
 & 0.2750 & 0.3035 & 0.2211

\\
\hline
\rowcolor{gray!20}\textbf{LMME3DHF (Ours)} 

& \bf\textcolor{red}{0.9000} & \bf\textcolor{red}{0.9415} & \bf\textcolor{red}{0.7363} 

& \bf\textcolor{red}{0.9203} & \bf\textcolor{red}{0.9555} & \bf\textcolor{red}{0.7694} 

\\
\rowcolor{gray!20}\textit{Improvement} & +6.6\% & +7.1\% & +8.9\% & +6.5\% & +8.2\% & +10.8\%
\\
\bottomrule
\end{tabular}}
\label{benchmark}
\end{table}
\begin{table}[h]
  \caption{Performance comparisons of LMMs on visual question-answering tasks. The best results are marked in \textcolor{red}{RED} and the second-best in \textcolor{blue}{BLUE}.}
  \label{qa_benchamrk}
  \centering
  \vspace{-3mm}
  \begin{tabular}{l|c}
    \toprule
    \textbf{Models} & \textbf{Accuracy (\%)}
    \\
    \midrule
    \textbf{VideoChat2 (7B)} \cite{li2023videochat}  & 16.63
    \\
    \textbf{InternVL2.5 (8B)} \cite{chen2024expanding} & \bf\textcolor{blue}{45.24}
    \\
    \textbf{MiniCPM-V2.6 (8B)} \cite{yao2024minicpm} & 13.83
    \\
    \textbf{Qwen2-VL (7B)} \cite{wang2024qwen2} & 23.18
    \\
    \textbf{Video-ChatGPT} \cite{maaz2023video} & 8.490
    \\
    \bottomrule
    \rowcolor{gray!20}\textbf{LMME3DHF (Ours)} & \bf\textcolor{red}{54.24}
    \\
    \bottomrule
  \end{tabular}
\end{table}
To evaluate the correlation between predicted scores and the ground-truth MOS, we employ three evaluation metrics: Spearman's Rank Correlation Coefficient (SRCC) \cite{spearman1904proof}, Pearson's Linear Correlation Coefficient (PLCC) \cite{pearson1896vii}, and Kendall's Rank Correlation Coefficient (KRCC) \cite{kendall1938new}. For assessing visual question answering performance, we use average accuracy as the evaluation metric. For the distortion-aware saliency prediction task, we adopt five commonly used consistency metrics: Area Under the Curve (AUC) \cite{hanley1982meaning}, Normalized Scanpath Saliency (NSS) \cite{peters2005components}, Correlation Coefficient (CC) \cite{pearson1896vii}, Similarity Metric (SIM) \cite{swain1991color}, and Kullback–Leibler Divergence (KLD) \cite{kullback1951information}. For deep learning-based IQA and VQA models, we partition the Gen3DHF database into training and test sets at a same ratio of \begin{math}4:1\end{math} as the previous literature. The models are implemented with PyTorch and trained on a 40GB NVIDIA RTX A6000 GPU with batch size of 8. We use the Adam optimizer with the initial learning rate set as \begin{math}1e{-}4\end{math} and set the batch size as 4. The training process is stopped after 50 epochs. For deep-learning based saliency predict models, we retrain them on the proposed Gen3DHF database using their officially realeased code and utilizing the same training, testing sets as mention above. The training parameters remain the same, but is conducted for only 30 epochs with an early stopping strategy applied. All experiments for each LMME3DHF are averaged using 5-fold cross-validation.

\begin{table}[t]
\vspace{-3mm}
  \caption{Performance comparisons of the state-of-the-art saliency models on distortion-aware saliency prediction. The best results are marked in \textcolor{red}{RED} and the second-best in \textcolor{blue}{BLUE}.}
  \label{saliency_benchmark}
  \centering
  \vspace{-3mm}
  \resizebox{0.47\textwidth}{!}{
  \begin{tabular}{l|ccccc}
    \toprule
    Model / Metric & AUC↑ & NSS↑ & CC↑ & SIM↑ & KLD↓
    \\
    \midrule
    \textbf{AIM} \cite{bruce2005saliency} & 0.6374 & 0.7306 & 0.0423 & 0.0163 & 4.6026
    \\
    \textbf{CA} \cite{goferman2011context} & 0.5893 & 0.5113 & 0.0297 & 0.0156 & 4.7227
    \\
    \textbf{CovSal} \cite{erdem2013visual} & 0.5306 & 0.0833 & 0.0051 & 0.0121 & 5.7113
    \\
    \textbf{GBVS} \cite{harel2006graph} & 0.5796 & 0.4507 & 0.0265 & 0.0154 & 4.7863
    \\
    \textbf{HFT} \cite{li2012visual} & 0.5728 & 0.3858 & 0.0223 & 0.0151 & 4.7867
    \\
    \textbf{IT} \cite{itti2002model} & 0.4969 & 0.0822 & 0.0049 & 0.0120 & 7.2030
    \\
    \textbf{Judd} \cite{judd2009learning} & 0.5058 & 0.1850 & 0.0097 & 0.0126 & 4.8111
    \\
    \textbf{Murray} \cite{murray2011saliency} & 0.6251 & 0.6491 & 0.0340 & 0.0146 & 4.6779
    \\
    \textbf{PFT} \cite{guo2008spatio} & 0.5624 & 0.3682 & 0.0185 & 0.0142 & 4.8277
    \\
    \textbf{SMVJ} \cite{cerf2007predicting} & 0.5785 & 0.4426 & 0.0261 & 0.0154 & 4.7884
    \\
    \textbf{SR} \cite{hou2007saliency} & 0.5253 & 0.2517 & 0.0144 & 0.0132 & 4.8328
    \\
    \textbf{SUN} \cite{zhang2008sun} & 0.6291 & 0.7432 & 0.0369 & 0.0156 & 4.6312
    \\
    \textbf{SWD} \cite{duan2011visual} & 0.4907 & 0.1450 & 0.0074 & 0.0125 & 4.9388
    \\
    \midrule
    \textbf{SALICON} \cite{huang2015salicon} & 0.6674 & 0.8193 & 0.0461 & 0.0171 & 4.5764
    \\
    \textbf{Sal-CFS-GAN} \cite{che2019gaze} & \bf\textcolor{blue}{0.7468} & \bf\textcolor{blue}{1.8207}
    & \bf\textcolor{blue}{0.1067} & \bf\textcolor{blue}{0.0441} & \bf\textcolor{blue}{4.1531}
    \\
    \textbf{TranSalNet-ResNet} \cite{lou2022transalnet} & 0.6655 & 0.9245 & 0.0520 & 0.0194 & 4.5461
    \\
    \textbf{TranSalNet-Dense} \cite{lou2022transalnet} & 0.6506 & 1.0633 & 0.0594 & 0.0213 & 4.5431
    \\
    \bottomrule
    \rowcolor{gray!20} \textbf{LMME3DHF (Ours)} & \bf\textcolor{red}{0.8388} & \bf\textcolor{red}{6.7689} 
    & \bf\textcolor{red}{0.5928} & \bf\textcolor{red}{0.4490} & \bf\textcolor{red}{1.8759}
    \\
    \bottomrule
  \end{tabular}}
\end{table}
\begin{table}[t]
\vspace{-2mm}
  \caption{Ablation study of the proposed LMME3DHF on distortion-aware saliency prediction performance.}
  \label{saliency_ablation}
  \centering
  \vspace{-3mm}
  \resizebox{0.47\textwidth}{!}{
  \begin{tabular}{cl|ccccc}
    \toprule
    No. & Training Strategy & AUC↑ & NSS↑ & CC↑ & SIM↑ & KLD↓
    \\
    \midrule
    (1) & w/o LLM & 0.6233 & 1.1929 & 0.0896 & 0.0716 & 6.3781
    \\
    (2) & Empty Prompt & 0.4574 & 0.0026 & 0.0025 & 0.0127 & 5.0163
    \\
    (3) & Only multimodal features & 0.6587 & 1.1822 & 0.0623 & 0.0203 & 4.4599
    \\
    \rowcolor{gray!20} (4) & \textbf{LMME3DHF (Ours)} & 0.8388 & 6.7689 & 0.5928 & 0.4490 & 1.8759
    \\
    \bottomrule
  \end{tabular}}
\vspace{-2mm}
\end{table}
\subsection{Performance on Score Prediction and Visual Question Answering}
As shown in Table \ref{benchmark}, traditional handcrafted IQA metrics such as BRISQUE \cite{mittal2012no} and NIQE \cite{mittal2012making} perform poorly in our scenario, indicating that their handcrafted features are primarily designed for natural image distortions and do not generalize well to AI-generated 3D human faces. On the other hand, although LLM-based metrics are widely recognized for their advanced visual understanding and visual question-answering capabilities, they fall short in accurately assessing perceptual video quality. In contrast, deep learning-based metrics, whether initially designed for IQA or VQA demonstrate significantly better performance compared to both handcrafted and LMM-based approaches. However, while these models achieve moderate to high performance in perceptual quality assessment, they generally lack visual question-answering capabilities, which are essential for improving the interpretability and diagnostic feedback in AI-generated content evaluation. Our proposed method LMME3DHF achieves the best performance from both quality and authenticity perspectives. This confirms the effectiveness of our model in evaluating human visual experience for AI-generated 3D HFs from multiple perspectives.

To assess the model performance on visual question answering, we further launch comparison of visual question-answering performance between our proposed LMME3DHF and various LMM-based metrics, as shown in Table \ref{qa_benchamrk}. Models are asked to identify and classify the types of distortions present in the video content. As the results show, LMME3DHF significantly outperforms all other baselines, highlighting its strength in both perceptual understanding and detailed diagnostic capabilities.

\subsection{Performance on Distortion-aware Saliency Prediction}
\begin{figure*}
    \centering
    \includegraphics[width=0.98\textwidth]{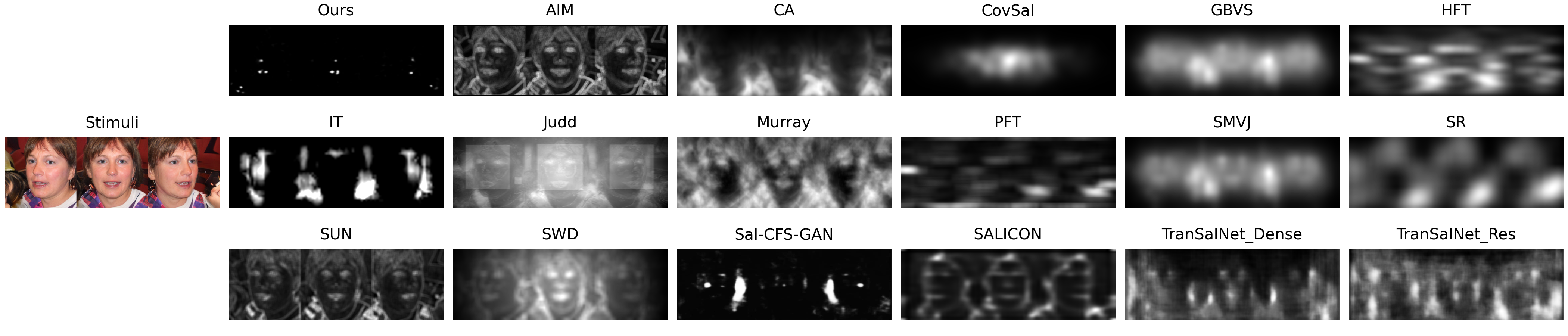}
    \vspace{-5mm}
    \caption{Qualitative comparisons for different models on our Gen3DHF.} 
     \vspace{-1mm}
    \label{saliency_exp}
\end{figure*}
\begin{figure*}
    \centering
    \includegraphics[width=0.9\textwidth]{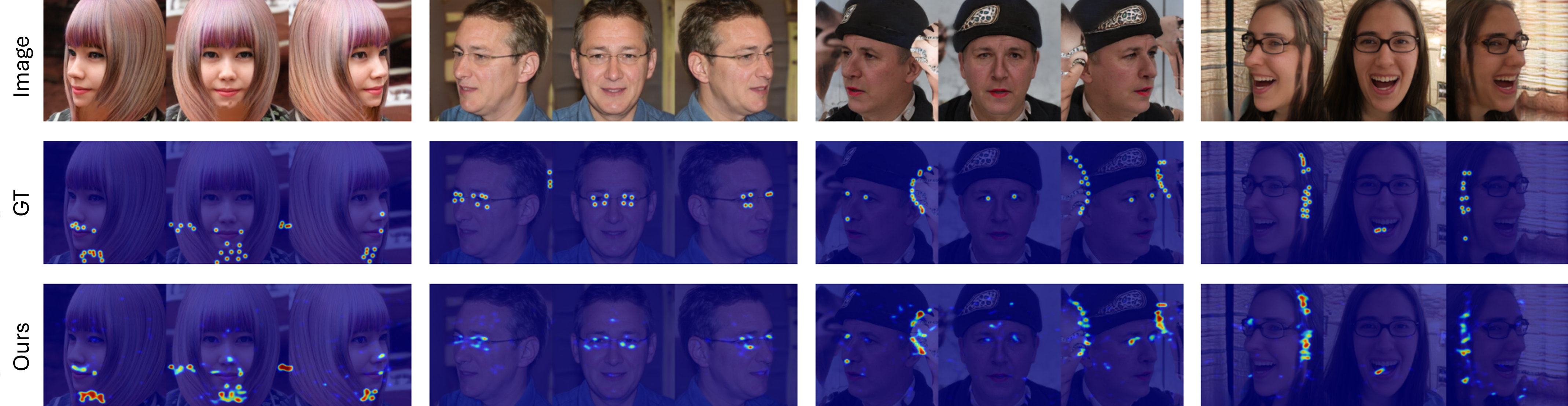}
    \vspace{-4mm}
    \caption{The predicted saliency maps of LMME3DHF on sample images from Gen3DHF dataset.} 
     \vspace{-1mm}
    \label{saliency_example}
\end{figure*}
\begin{table*}[h]
  \caption{Ablation study of the proposed LMME3DHF on score prediction and visual question answering performance.}
  \vspace{-3mm}
  \renewcommand\arraystretch{0.95}
  \resizebox{0.85\textwidth}{!}{
  \begin{tabular}{cccc|ccc|ccc|c}
    \toprule
    & \multicolumn{3}{c}{Training Strategy} & \multicolumn{3}{c}{Quality} 
    & \multicolumn{3}{c}{Authenticity} & \multicolumn{1}{c}{Accuracy}
    \\
    \cmidrule(r){2-4} \cmidrule(r){5-7} \cmidrule(r){8-10} \cmidrule(r){11-11}
    \multicolumn{1}{c}{No.} & LoRA$_{r=8}$ (vision) & LoRA$_{r=8}$ (LLM) & Quality Regression
    & SRCC & PLCC & KRCC & SRCC & PLCC & KRCC & (\%)
    \\
    \hline
    \multicolumn{1}{c}{(1)} & \ding{52} & & 
    & 0.0324 & 0.0277 & 0.0251 & 0.0316 & 0.0325 & 0.0258 & 30.86
    \\
    \multicolumn{1}{c}{(2)} & & \ding{52} & 
    & 0.7000 & 0.7573 & 0.5807 & 0.7137 & 0.7627 & 0.5923 & 44.83
    \\
    \multicolumn{1}{c}{(3)} & & & \ding{52}
    & 0.8589 & 0.9345 & 0.6715 & 0.8841 & 0.9469 & 0.7111 & 30.73
    \\
    \multicolumn{1}{c}{(4)} & \ding{52} & \ding{52} & 
    & 0.6976 & 0.7517 & 0.5775 & 0.6917 & 0.7431 & 0.5736 & 41.57
    \\
    \rowcolor{gray!20} \multicolumn{1}{c}{(5)} & & \ding{52} & \ding{52}
    & 0.9028 & 0.9424 & 0.7406 & 0.9234 & 0.9571 & 0.7741 & 54.24
    \\
    \multicolumn{1}{c}{(6)} & \ding{52} & \ding{52} & \ding{52}
    & 0.9052 & 0.9420 & 0.7470 & 0.9226 & 0.9561 & 0.7731 & 53.12
    \\
    \bottomrule
  \end{tabular}
  \label{ablation}}
  \centering
\end{table*}
To evaluate performance on the distortion-aware saliency prediction task, we compare LMME3DHF with current state-of-the-art saliency prediction models, including both traditional and deep learning-based approaches. As shown in Table \ref{saliency_benchmark}, LMME3DHF significantly outperforms all baseline models across various evaluation metrics. Unlike general saliency detection that focuses on broad visual attention, our task targets sparse, distortion-aware saliency, identifying precise regions associated with visual distortions. As a result, commonly used saliency prediction models are not well-suited for this specialized task, leading to limited performance in comparison. The visual comparisons presented in Figure \ref{saliency_exp} and the predictions on sample images from the Gen3DHF dataset in Figure \ref{saliency_example} further emphasize that LMME3DHF outperforms other saliency prediction models, clearly demonstrating its superior ability to accurately localize distortion-aware salient regions.

\subsection{Ablation Study}
We conduct ablation experiments to validate the contributions of the key components in our proposed LMME3DHF framework. Table \ref{saliency_ablation} shows the results for saliency-distortion-aware decoder. Experiments (1) show that removing features from the language decoder causes a noticeable drop in performance. Experiment (2) further emphasizes this scenario by setting an empty prompt, thereby demonstrating the critical role of textual guidance. Experiment (3) shows that using only fused multimodal features enhances the model’s ability to predict distortion-aware saliency. Experiment (4) our approach that combines both visual and multimodal features achieves the highest performance, confirming the effectiveness of integrating both feature types. Results for score prediction and visual question answering are summarized in Table \ref{ablation}. Experiment (1) which fine-tuning only the vision encoder, yields the weakest performance across both tasks. Experiment (2) demonstrates that fine-tuning the LLM significantly improves performance, with the improvement being especially pronounced in the visual question answering task, which is inherently text-based. Experiment (3) shows a significant improvement in score prediction tasks, highlighting the effectiveness of the quality regression module. Experiment (4) performs comparably to Experiment (2), suggesting that the vision encoder is less critical in improving evaluation results. Finally, Experiments (5) and (6) achieve the best overall performance. Among them, we select the configuration from Experiment (5) for LMME3DHF, as it balances strong performance with computational efficiency.

\section{CONCLUSION}
In this paper, we investigate the problem of human visual preference evaluation for AI-generated 3D HFs. We introduce Gen3DHF, which includes 2,000 videos of 3D HFs generated by five different models, evaluated from both quality and authenticity dimensions and annotated with MOSs and distortion mark-description pairs. Using Gen3DHF, we evaluate state-of-the-art quality assessment models and establish a new benchmark for this task. Building upon this dataset, we propose LMME3DHF, an LMM-based evaluation model that leverages instruction tuning and LoRA techniques to perform perceptual quality assessment and predict distortion-aware saliency along with descriptive explanations. Extensive experiments demonstrate that LMME3DHF achieves state-of-the-art performance for quality evaluation and distortion-aware saliency prediction tasks for Gen3DHF. We hope that LMME3DHF will serve as a valuable tool for advancing research in the generation and evaluation of AI-generated 3D human faces.

\section*{acknowledgements}
This work was supported in part by the National Natural Science Foundation of China under Grant 62271312 and Grant 62132006, and in part by STCSM under Grant 22DZ2229005.

\bibliographystyle{ACM-Reference-Format}
\bibliography{main}

\end{document}